\documentclass[journal]{IEEEtran}

\usepackage{color,array}

\usepackage[pdftex]{graphicx}
\DeclareGraphicsExtensions{.pdf,.jpeg,.png}

\usepackage[switch]{lineno} 
\usepackage{amsmath}
\usepackage{amssymb}
\usepackage{textcomp}
\usepackage{gensymb}
\usepackage{multirow}
\usepackage{bm}
\usepackage{graphicx}
\usepackage{tabularx}
\usepackage{xcolor}
\usepackage{soul}

\usepackage{booktabs}
\usepackage[ruled,vlined]{algorithm2e}

\newcommand{\etal}{\textit{et al}., }
\newcommand{\ie}{\textit{i}.\textit{e}., }
\newcommand{\eg}{\textit{e}.\textit{g}., }

\usepackage[colorlinks,urlcolor=blue,linkcolor=blue,citecolor=blue]{hyperref}

\usepackage[capitalize]{cleveref}

\usepackage{cite}

\begin{document}

\title{A Cyber Manufacturing IoT System for Adaptive Machine Learning Model Deployment by Interactive Causality Enabled Self-Labeling}

\author{Yutian Ren, Yuqi He, Xuyin Zhang, Aaron Yen, and  G. P. Li
\thanks{The authors are with the California Institute for Telecommunications and Information Technology, University of California, Irvine, CA 92697
USA (e-mail: yutianr@uci.edu; gpli@calit2.uci.edu).}
}


\maketitle

\begin{abstract}
Machine Learning (ML) has been demonstrated to improve productivity in many manufacturing applications. To host these ML applications, several software and Industrial Internet of Things (IIoT) systems have been proposed for manufacturing applications to deploy ML applications and provide real-time intelligence. Recently, an interactive causality enabled self-labeling method has been proposed to advance adaptive ML applications in cyber-physical systems, especially manufacturing, by automatically adapting and personalizing ML models after deployment to counter data distribution shifts. The unique features of the self-labeling method require a novel software system to support dynamism at various levels.

This paper proposes the AdaptIoT system, comprised of an end-to-end data streaming pipeline, ML service integration, and an automated self-labeling service. The self-labeling service consists of causal knowledge bases and automated full-cycle self-labeling workflows to adapt multiple ML models simultaneously. AdaptIoT employs a containerized microservice architecture to deliver a scalable and portable solution for small and medium-sized manufacturers. A field demonstration of a self-labeling adaptive ML application is conducted with a makerspace and shows reliable performance.

\end{abstract}


\section{Introduction}
\label{sec:intro}

The integration of real-time machine learning (ML) technology into cyber-physical systems (CPS), such as smart manufacturing, requires a hardware and software platform to orchestrate sensor data streams, ML application deployments, and data visualization to provide actionable intelligence. Contemporary manufacturing systems leverage advanced cyber technologies such as Internet of Things (IoT) \cite{lu2016internet} systems, service-oriented architectures \cite{ser_ori1}, microservices \cite{cps_microservice}, and data lakes and warehouses \cite{rudack2022towards}. ML applications can be integrated with existing tools to support and enable smart manufacturing systems.
For example, Yen \etal developed a software-as-a-service (SaaS) framework for managing manufacturing system health with IoT sensor integration that can facilitate data and knowledge sharing \cite{yenSaaS}.
Mourtzis \etal proposed an IIoT system for small and medium-sized manufacturers (SMMs) incorporating big data software engineering technologies to process generation and transmission of data at the terabyte-scale monthly for a shop floor with 100 machines \cite{MOURTZIS2016290}.
Liu \etal designed a service-oriented IIoT gateway and data schemas for just-enough information capture to facilitate efficient data management and transmission in a cloud manufacturing paradigm \cite{LIU2022102217}.
Sheng \etal proposed a multimodal ML-based quality check for CNC machines deployed using edge (sensor data acquisition) to cloud (Deep Learning compute) collaboration \cite{SHENG2024102324}. 
Morariu \etal designed an end-to-end big data software architecture for predictive scheduling in service-oriented cloud manufacturing systems \cite{MORARIU2020103244}.
Paleyes \etal \cite{ml_deploy_challenge} summarized the challenges in deploying machine learning systems in each stage of the ML lifecycle. For manufacturing companies especially SMMs, the relatively outdated IT infrastructure, lack of IT expertise, and heterogeneous nature of manufacturing software and hardware systems complicate ML application deployment \cite{davis2020cyberinfrastructure}. 
While systems in the literature have demonstrated various ML applications, they lack support for adaptive ML.

A major component of the cyber manufacturing paradigm is actionable intelligence, providing users with critical information to act at the right time and place. Manufacturers significantly favor personalized intelligence for its ability to adapt to their specific use cases. However, barriers exist to the development and deployment of personalized ML systems in manufacturing environments. The cost of manually collecting and annotating a training dataset slows the democratization of ML-enhanced smart manufacturing systems, especially in SMMs \cite{davis2020cyberinfrastructure}. Recently, the development of adaptive machine learning, which autonomously adapts ML models to diverse deployment environments, has become a viable solution to lower the entry barrier to ML for SMMs. 
Several types of adaptive ML methods, including pseudo-labels empowered by semi-supervised learning (SSL) \cite{yan2021augmented, slb_neurips}, delayed labels \cite{delay2, delayed_lb_review}, and domain knowledge enabled learning \cite{domainknowledge, stewart2017label}, have been proposed.

A novel interactive causality based self-labeling method has been proposed to achieve adaptive machine learning and has been demonstrated in manufacturing cyber-physical system applications \cite{ren2023slb, tii}. This method utilizes causal relationships extracted from domain knowledge to enable an automatic post-deployment self-labeling workflow to adapt ML models to local environments. The self-labeling method works in real time to automatically capture and label data and is able to effectively utilize limited pre-allocated or public datasets.
Self-labeling is a coordinated effort between three types of computational models, namely task models, effect state detectors (ESDs), and interaction time models (ITMs), to execute the self-labeling workflow for adapting task models after deployment. An overview of this method is provided in \cref{sec:bg}. 
The merit of the self-labeling method is in its ability to fully leverage the unique properties of ML applications in CPS contexts, including scenarios with rich domain knowledge, dynamic environments with time-series data and possible data shifts, and diverse environments with limited pre-allocated datasets to fulfill the needs of personalized solutions at the edge.

To support and execute the interactive causality based self-labeling (SLB) method, especially for SMMs, the system infrastructure must support the following requirements:
1) real time timestamped data transfer of sensor, audio, and video data from from heterogeneous services and devices;
2) a causality knowledge base that manages the interaction between models to facilitate self-labeled ML between causally related nodes.
3) a core self-labeling service that connects the ML services, routes data streams, executes the self-labeling workflow, and retrains and redeploys ML models autonomously at the edge;
4) a scalable architecture to easily accommodate new edge, ML, and SLB services.
Due to the unique needs of interactive causality, a novel software system is required to realize self-labeling functionality for various ML models. This software system harnesses real-time IoT sensor data, ML, and self-labeling services to enable self-labeling adaption of models to ever-changing environments.

In this paper, we propose and implement the AdaptIoT system as a platform to develop cyber manufacturing applications with adaptive ML capability. The AdaptIoT platform employs mainstream software engineering practices to achieve an affordable, scalable, actionable, and portable (ASAP) solution for SMMs. 
AdaptIoT defines an end-to-end IoT data streaming pipeline that supports high throughput ($\geq$ 100k msg/s) and low latency ($\leq$ 1s) sensor data streaming via HTTP and defines a standard interface to integrate ML applications that ingest sensor data streams for inference. The most important feature of AdaptIoT is its inherent support for self-labeling, managing various computational models (\eg ML models) to automatically execute flexible self-labeling workflow to collect and annotate data without human intervention to retrain and redeploy ML models. A causality knowledge base is incorporated to store and manage the virtual interactions among computational models for self-labeling.
AdaptIoT employs a scalable micro-service architecture that can easily integrate future capabilities such as data shift monitoring.
We deploy AdaptIoT in a small-scale makerspace to simulate its application in SMMs and develop a self-labeling application using the AdaptIoT platform, demonstrating the its applicability and the adaptive ML capability of AdaptIoT in real-world environments. Part of the platform source code is open-sourced at \url{https://github.com/yuk-kei/iot-platform-backend}.

\section{Overview of Interactive Causality and Self-Labeling Method}
\label{sec:bg}

The Interactive Causality enabled self-labeling (SLB) method is developed to achieve fully automatic post-deployment adaptive learning for ML systems such that deployed ML models can adapt to local data distribution changes (\eg concept drift \cite{conceptdrift}.)  This section includes a brief review of the self-labeling technique.

\begin{figure}[!t]
\centering
\includegraphics[width=1.0\columnwidth]{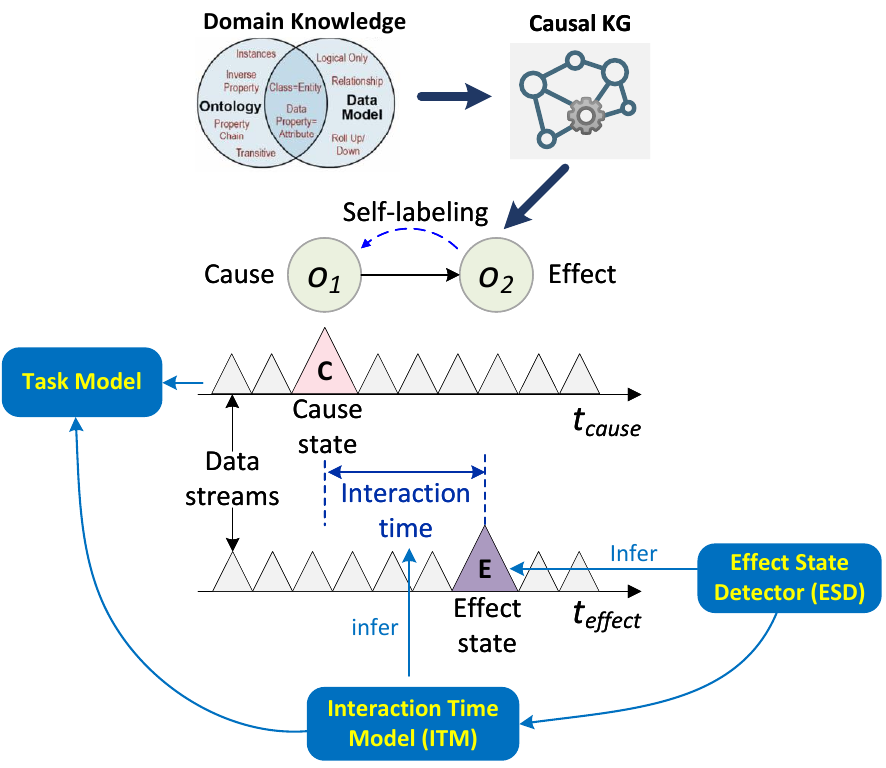}
\caption{An illustration of the overall procedure of self-labeling.}
\label{fig_slb}
\end{figure}

Self-labeling begins with selecting two causally connected nodes within a dynamic causal knowledge graph (KG) \cite{ckg}, which can be obtained from domain knowledge and ontology. In the minimum case where the selected nodes are adjacent, the cause and effect events are related by an interaction time between their occurrence. This interaction time can vary but typically has a correlation with the effect state transient \cite{ren2023slb}.

SLB requires monitoring one or more data streams so that the cause and effect state transitions can be observed. In \cref{fig_slb}, the cause and effect time-series data streams are collected at nodes $o_{1}$ and $o_{2}$ respectively. The first of the three models required for SLB is the effect state detector (ESD), which monitors the data streams that provide effect data and is responsible for identifying effect state transients, including classification. The interaction time model (ITM) intakes the effect data within windows selected by the ESD, optionally including the ESD output, and predicts the interaction time (\ie causal time lag \cite{timelag}) between the cause and effect state transitions. The relevant portion of the cause data stream is extracted using the effect state transition timestamp determined by the ESD and the interaction time output by ITM. With the cause data associated with effect state transitions, we use effect transitions as the label and cause data as the input features to train the task model.

The task model is our primary decision model, enriched by SLB through continual learning. Continual learning through self-labeling is particularly beneficial in scenarios where the input and/or output data distributions shift from their values during initial training. The relationship between cause and effect is resilient to drifts in data, and
this resiliency is inherited by the self-labeling method to provide a basis for continual learning. Time-series data streams are collected for each system, with the causal relationship defining a cause and an effect system. A key advantage of the self-labeling method is its ability to independently detect and label the effect system and propagate said label to the relevant time-series data in the cause system, automating the continual learning described above. This allows for a robust predictive classifier to be implemented without necessitating human intervention to facilitate continual learning.

\section{Software Architecture of AdaptIoT for Self-Labeling}
\label{sec: multi}

In this section, we describe AdaptIoT's modular software architecture and illustrate the specialized modules for self-labeling applications.

\begin{figure*}[!t]
\centering
\includegraphics[width=0.7\textwidth]{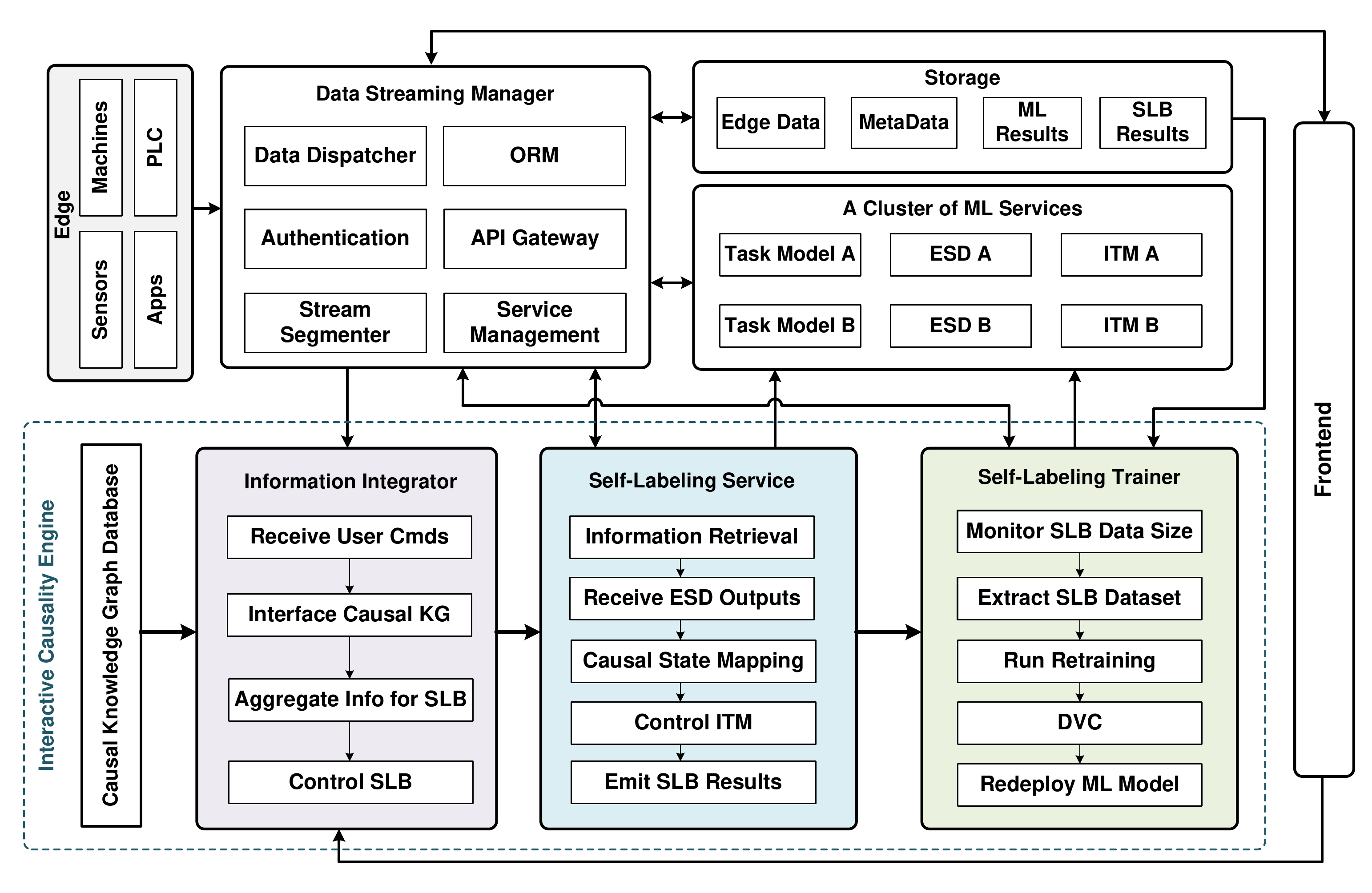}
\caption{A high-level block diagram of the proposed IIoT system for self-labeling applications.}
\label{fig_block}
\end{figure*}

\subsection{Module-level Architecture}

To meet the unique requirements of self-labeling applications as stated in \cref{sec:intro}, a high-level system block diagram is illustrated in \cref{fig_block}. The major functional modules of the system are composed of edge services, the Data Streaming Manager (DSM), databases for storage, and clusters of ML services (note that even though named as ML services for generalizability, some of them can be simple data/signal processing or statistical models), the Interaction Causality Engine (ICE), and a frontend Graphical User Interface (GUI) handler. The edge services comprise of sensors, edge computing devices, external applications, and machines on the factory floor. The \textit{in situ} edge services stream data via the DSM to databases and applications, including the ICE and ML services. 
The DSM is the main message broker, routing high-throughput streaming data generated by edge, ML, and ICE services to their destination. The DSM serves as the backend for data streaming and service management.

To efficiently store various types of data, multiple database types are implemented, including time-series databases, SQL, and no-SQL databases. The databases store raw timestamped sensor data, metadata for services and devices, processed ML results, and self-labeling results. In addition, a cluster of ML services, including task models, ESDs, and ITMs, runs to provide actionable intelligence while participating in the self-labeling workflow.

\subsection{Interactive Causality Engine}
The Interactive Causality Engine (ICE) is the core engine enabling adaptability for deployed ML task models. ICE consists of a causal knowledge graph database, an information integrator, a self-labeling service, and a self-labeling trainer. The four components undertake different tasks and jointly execute the self-labeling workflow in an automatic manner. 

\begin{figure}[!t]
\centering
\includegraphics[width=1.0\columnwidth]{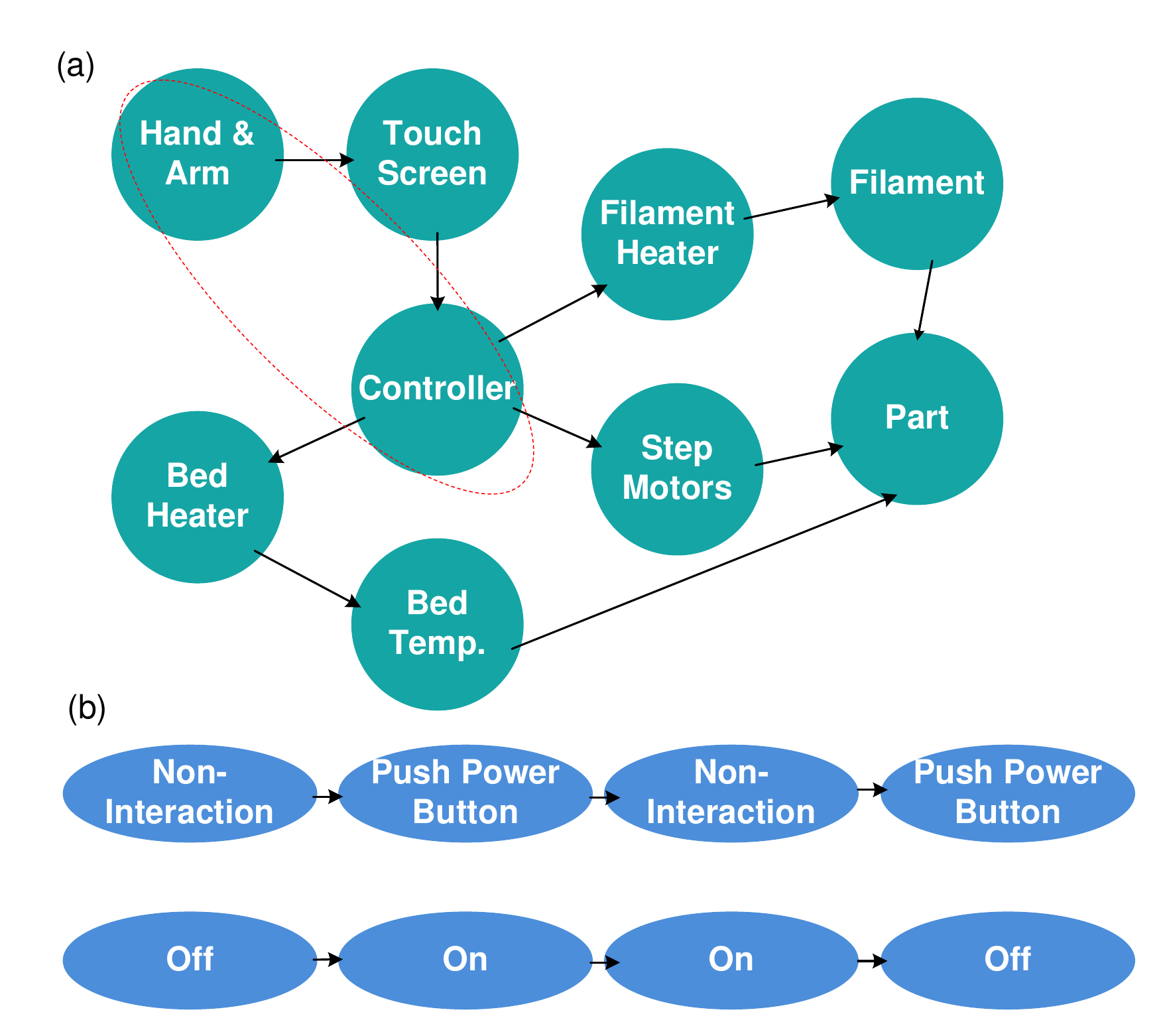}
\caption{(a) A simplified knowledge graph for a 3D printer use case; (b) Corresponding state transition relationships of the causally related Hand\&Arm node and Controller node.}
\label{fig_kg}
\end{figure}

The causal knowledge graph database stores multiple KGs with directional links that represent the interactivity and underlying causality among the linked nodes. These KGs are extracted and reformulated from existing domain knowledge. A simplified KG sample of a 3D printer is shown in \cref{fig_kg}(a). A direct link in this graph represents interactivity, and connections between nodes suggest the possibility of causality between the nodes. The links in the KG can be bidirectional, differing from many causal graph model definitions (such as structural causal models \cite{pearl2009causality}) as the low-level causal relationships between connected nodes are broken down into state-level representations in the temporal scale. In simple terms, at a high level two bilaterally linked nodes can be mutually causally related but at finer temporal resolutions, in any instance one side serves exclusively as the cause and the other as the effect.
Given two connected nodes, the state transition mappings (\ie logical relations) of two nodes are represented as a dynamic and temporal state machine as exemplified in \cref{fig_kg}(b). We represent this state machine with corresponding state transition relationships by using a truth table.

The information integrator bridges the causal KG database, the self-labeling service, sensor metadata, ML services, and users to ingest and integrate needed information and control the self-labeling. Through the information integrator, users can start or stop a self-labeling workflow among the causally linked nodes. The information integrator also scrutinizes the information completeness for running a self-labeling service.

The self-labeling service receives inputs from the information integrator and initiates a self-labeling workflow by coordinating the raw data streams from sensors, corresponding ML services, and the self-labeling trainer. When a self-labeling service starts, the following functions will be executed: 
1) receive control signals from the information integrator to start or stop a self-labeling workflow; 
2) receive inputs from the information integrator, including the selected causal nodes, the truth table representing the causal logical relations, the URLs of corresponding sensor streams and ML services, and the output paths (URLs); 
3) receive outputs from ESDs and execute causal state mapping to find consistent cause states;
4) assemble inputs for ITMs and route them to corresponding ITMs; 
5) receive ITM outputs, combine ITM outputs with corresponding cause states, and emit them to a database for storage; 
6) optionally select corresponding data segments from cause streams based on the information in Step 5. 
Note that since the actual interaction time of each effect can be very different, Step 3 inspects whether to self-label the causes needs to wait additional effect states being detected. The self-labeling service can run multiple self-labeling workflows in parallel for various nodes in KGs.

The self-labeling trainer is an independent and decoupled service that constantly monitors the number of self-labeled samples, receives users' commands via the information integrator, retrains task models, and redeploys the task models. It is designed to be separate from the self-labeling service for reusability and extensibility. The self-labeling trainer will schedule a training session at non-peak hours when the number of self-labeled samples reaches the requirements with user approval. In addition, data version control (DVC) \cite{mlops} is applied to version self-labeled datasets and trained weight files for MLOps to efficiently manage the continuous retraining empowered by self-labeling. After retraining, users can choose whether to redeploy the task model with the new weights.

\subsection{Unit Service Model}

To connect and scale to heterogeneous edge services and ML services, an abstract layer-wise unit service model is designed to work as the fundamental architecture for a single service in the proposed AdaptIoT system. The unit service model is designed to accommodate and standardize all types of services in the system that generates data and sends generated data to a storage place. This layer-wise architecture for a single service ensures the scalability and homogeneity of downstream interfaces. The unit service model is abstracted into four layers from the bottom up: the asset layer, data generation (DataGen) layer, service layer, and API layer.

Asset layer defines an abstraction of the independent components connected to the system, such as hardware (\eg sensors and machines), external applications (\eg proprietary software), or external data sources (\eg external database). A key uniqueness for this layer is that the system can interface with the independent components to receive data or run applications but cannot control or access their sources.

The data generation layer encapsulates a software that generates one data sample upon called once. This layer performs the core function of data generation by interacting with the Asset layer. A higher level abstraction of heterogeneous edge applications is achieved in this layer by defining uniform Class attributes and functions. For example, the sensor firmware as the asset layer communicates with the DataGen layer to retrieve one data sample per call. The inference function of a ML model using various ML frameworks, \eg scikit-learn, PyTorch, or Tensorflow, is unified with the same interface to interact with the Service layer. 
To receive data generated by external applications, we define a $Receiver$ function using REST API to accept a POST request from external applications and data sources. The POST request after scrutinization is rerouted in the $Receiver$ for the DataGen layer to use GET to acquire samples individually. 

The Service layer integrates necessary functions as a microservice on top of the data generation. It handles receiving inputs from upper API layers, \ie inputs needed for ML inference in the DataGen layer. It integrates the inputs and the DataGen layer to generate data in a discrete or continuous way. Upon new data is generated, an emitter function is executed to send out data to the following pipeline. Besides interacting with the DataGen layer, the Service layer integrates other auxiliary functions, including control (\ie start, stop, update), service registration, and metadata management, for the API endpoints in the API layer. Up to this layer, all the heterogeneous applications from the bottom are consolidated with a homogeneous interface.

The top layer is the API layer, where the API endpoints are defined using a web framework. This layer handles all the API-level I/O and interactions with other services by calling functions defined in the Service layer.

Besides the four basic layers, an orchestration layer is designed to moderate the same type of services with same or different configurations operating on the same hardware. This layer is optional, depending on the actual needs of service orchestration.

\section{System Implementation and Analysis}
\label{sec: imple}

This section details the system implementation of AdaptIoT, including the software and hardware infrastructure, and provides an example implementation of a self-labeling service hosted on AdaptIoT.

\begin{figure*}[!t]
\centering
\includegraphics[width=1.0\textwidth]{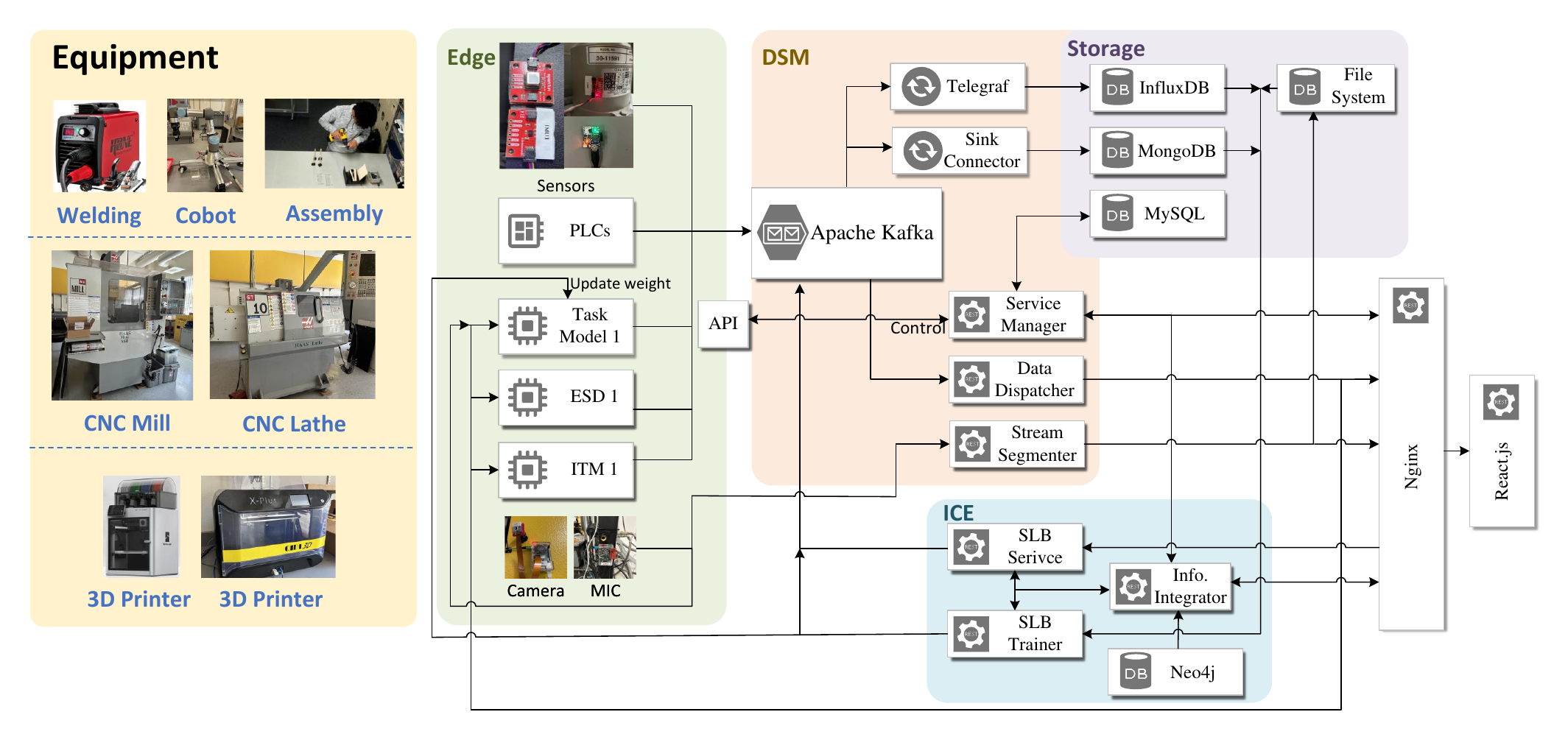}
\caption{Hardware and software infrastructure. Each block represents a containerized software service.}
\label{fig_arch}
\end{figure*}

\begin{figure}[!t]
\centering
\includegraphics[width=\columnwidth]{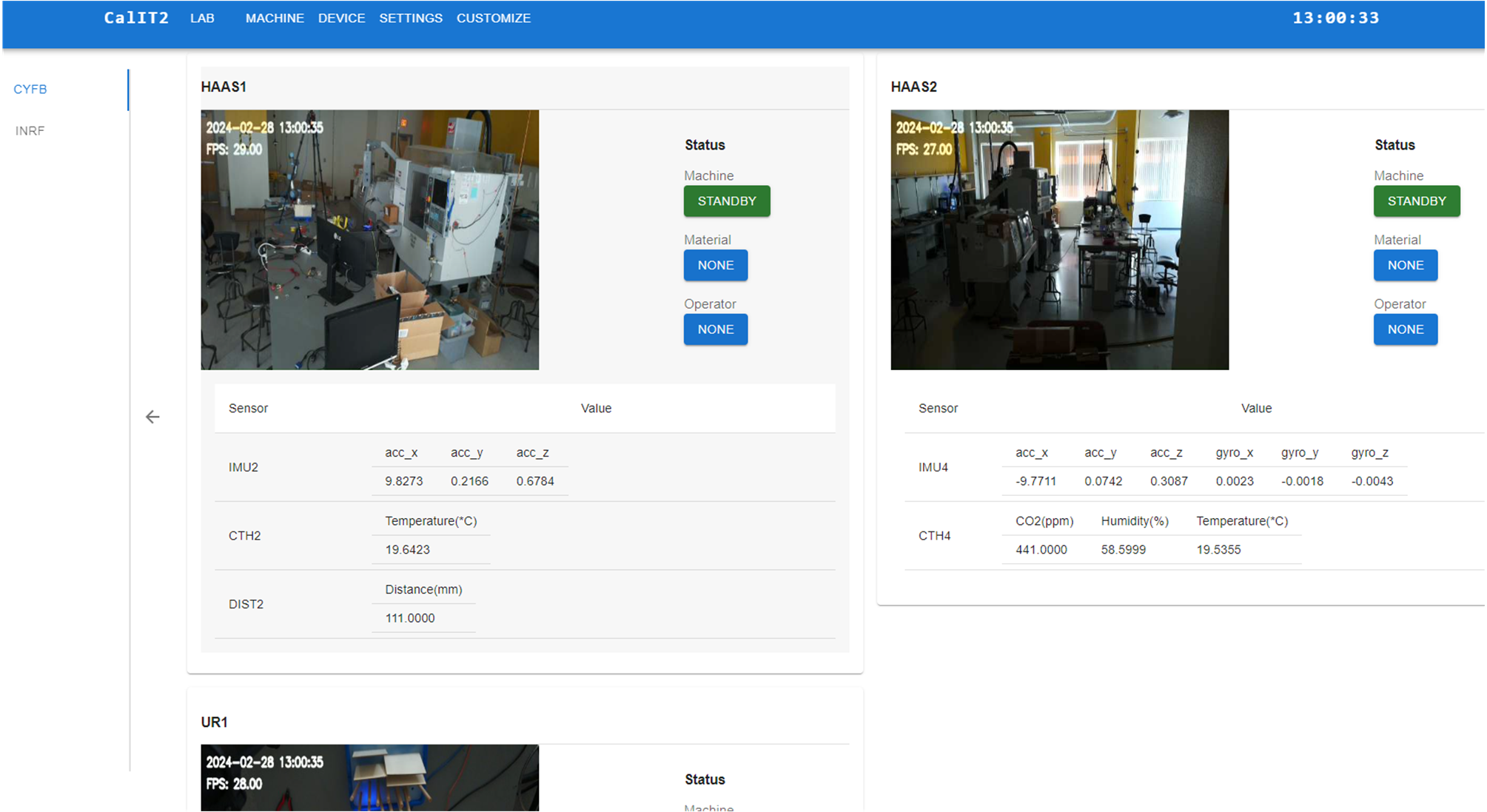}
\caption{The web GUI to display real-time data and ML results where users can also apply control to the system.}
\label{fig_gui}
\end{figure}

\subsection{Hardware Infrastructure of the Cyber MakerSpace}
\cref{fig_arch} illustrates a complete implementation of the proposed AdaptIoT system for self-labeling applications and the hardware infrastructure, including manufacturing equipment. This system is deployed to a cyber makerspace lab with common manufacturing equipment including 3D printers, CNC Machines (Mill and Lathe), a collaborative robot, and TIG welding machines. 
We heavily instrument each machine and the entire space with sensors to collect multimodal signals, including cameras, power meters, vibration, acoustic, distance, environmental, and other specialized sensors. The sensors are installed at multiple locations and critical components of a machine. In addition, the CNC machines and the robot are controlled by programmable logic controllers (PLCs) that are interfaced to acquire information about machine running status directly.

\subsection{AdaptIoT System Implementation}
The implemented services and software components are shown in \cref{fig_arch} with examples of edge services. \cref{fig_gui} shows an example of the web-based GUI. Each block in \cref{fig_arch} represents an independent dockerized \cite{merkel2014docker} web service running on various hardware and executes one or more functions as described in \cref{fig_block}. Except for databases, all other communications among services are via REST API by the lightweight Flask web framework. Four services, service manager, data dispatcher, stream segmenter, and information integrator, are exposed to the React frontend via Nginx. The internal APIs are hidden behind the four previously mentioned services.

\subsubsection{Software Components}

\textbf{Message queue.} A message queue is a communication method used in distributed systems and computer networks for asynchronous communication between various components or processes. The key feature of a message queue is that it decouples the producers and consumers in terms of time and space. Producers and consumers do not need to run simultaneously or on the same machine. This decoupling is useful in building scalable and flexible systems, as components can communicate without being directly aware of each other. Due to these features, we choose a message queue as the main message broker in DSM. Popular message queue systems include Apache Kafka, RabbitMQ, and Apache ActiveMQ \cite{dobbelaere2017kafka, mq2}. 
This study uses Kafka due to its outstanding horizontal scalability and high throughput.

\textbf{Database and storage.}
Several types of data are needed to be stored, and accordingly, several types of storage are chosen. We consider the factors including data structure, throughput, size, access frequency, and scalability. A MySQL database stores static metadata for all the services and users. For example, the relational metadata for a sensor service includes its factory locations, associated machines, vendor information, and URL for getting data. An IoT system with streaming sensors requires a continuous high data throughput (\eg $\geq$ 10k samples/sec), which puts additional demand on database ingestion speed. Time series databases are typically designed to handle high throughput, especially in scenarios with a continuous influx of timestamped data. For storing high-throughput sensor data, a time-series database InfluxDB \cite{influxdb} is chosen.
Regarding the results generated by ML services, we use both MongoDB and MySQL, depending on the data types. In addition, a graph database Neo4j \cite{neo4j} is chosen to store the causal knowledge graph. Video and audio data are stored in file systems only.

\begin{figure}[!t]
\centering
\includegraphics[width=1.0\columnwidth]{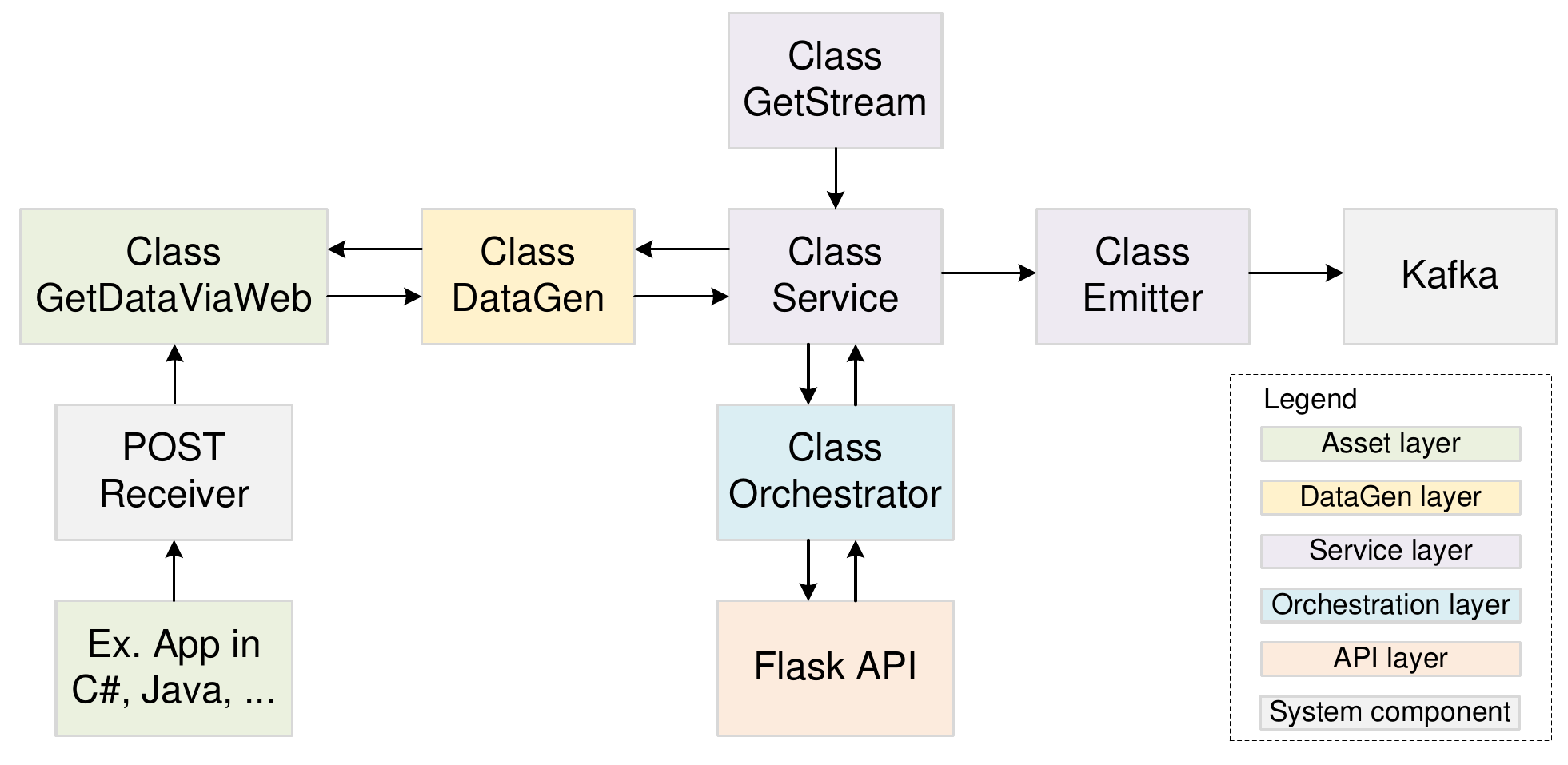}
\caption{A unit service example of receiving data from an external application.}
\label{fig_ser_eg}
\end{figure}

\textbf{Implementation of the unit service model.} 
\cref{fig_ser_eg} describes the detailed implementation of a unit service model in connecting with an external application. 
Starting from the asset layer, external applications in various languages and platforms can post an authenticated JSON format message via REST API to the $Receiver$ from which an asset layer function \textit{GetDataViaWeb} can get data. The $Receiver$ connects with the API gateway to serve as the system-level data ingestion service for external data sources.
The \textit{DataGen} class calls \textit{GetDataViaWeb} continuously to acquire the JSON message and wrap it into a defined standard message format. The \textit{Service} class in the service layer integrates \textit{DataGen} and input sources \textit{GetStream} to execute service-level functions. The received inputs from \textit{GetStream} in JSON format are transmitted downwards to the data generation layer and possibly asset layer for processing. In class \textit{Service}, the generated data is sent to an \textit{Emitter} that manages all the data transmission to Kafka. An \textit{Orchestrator} stays on top of multiple \textit{Service} of same type for control and interaction with other unit services. A Flask API layer provides a lightweight web server for each unit service and defines the API endpoints on the top.

\subsubsection{Data Flow}
As an illustration, we will describe a complete data flow in AdaptIoT from an edge sensor to an ML service. An edge sensor encapsulated in a unit service model generates a sample and emits it to the Kafka cluster. In the Kafka cluster, the sample is allocated to a partition for processing, after which this sample is routed to two places. First, the sample is routed by Telegraf to the InfluxDB for persistence. In the meantime, due to the unique requirement of many ML applications that need continuous data processing, the \textbf{Data Dispatcher} is implemented to route received individual samples into an HTTP data streaming via Server Sent Events (SSE) \cite{sse} and a query interface via REST API. ML services that need this data stream can use the standard HTTP method to receive the stream. 
The inferred ML results are emitted to Kafka again and routed to the corresponding MongoDB and data dispatcher. The React frontend queries the APIs for visualization.

\subsection{ICE implementation.}
Two types of data structures are used to represent the causality among nodes in a KG and the exact causal logical relations between any selected nodes. For the causal knowledge graph, we use a graph database Neo4j \cite{neo4j} to represent the nodes, attributes of nodes, and the directional relationships among nodes. The truth table is used to represent various causal logical relations among arbitrary nodes. The truth tables are stored in a MongoDB in key-value pairs.

We define a standard class $SlbService$ that can apply the self-labeling method on any causally related ML services given relevant parameters. The outputs of self-labeling are three key values by fusing the outputs of ESD and ITM, including a corresponding cause state, a timestamp of the end of the cause state, and the duration of the cause state. To partition the cause data streams based on self-labeling results, the system supports operation in two modes. Mode 1 saves the raw self-labeling outputs in MongoDB that are used to generate a retraining dataset by the SLB trainer afterward. Mode 2 is to create self-labeled data samples on the fly when $SlbService$ is running to provide immediate feedback for users. Both modes can be turned on at the same time. 
The SLB trainer independently monitors the number of self-labeled samples by querying the database at a constant frequency and manages ML training scripts for retraining ML models.

\textbf{Negative Samples.} Similar to other natural label-based systems, \ie social media recommendation systems \cite{covington2016deep} where users' interactions (likes, views, comments) are used as positive labels, in many cases, the ESD can only provide positive labels when there are state transitions different from the background distribution. The acquisition of negative samples from the background data distribution follows the same strategy as recommendation systems via negative sampling or more advanced importance sampling. The negative sampling is undertaken by each ESD since ESD keeps a buffer of its own historical states. The ESD randomly samples the background distribution as the negative labels and sends them to the self-labeling service for processing.

\begin{figure}[!t]
\centering
\includegraphics[width=1.0\columnwidth]{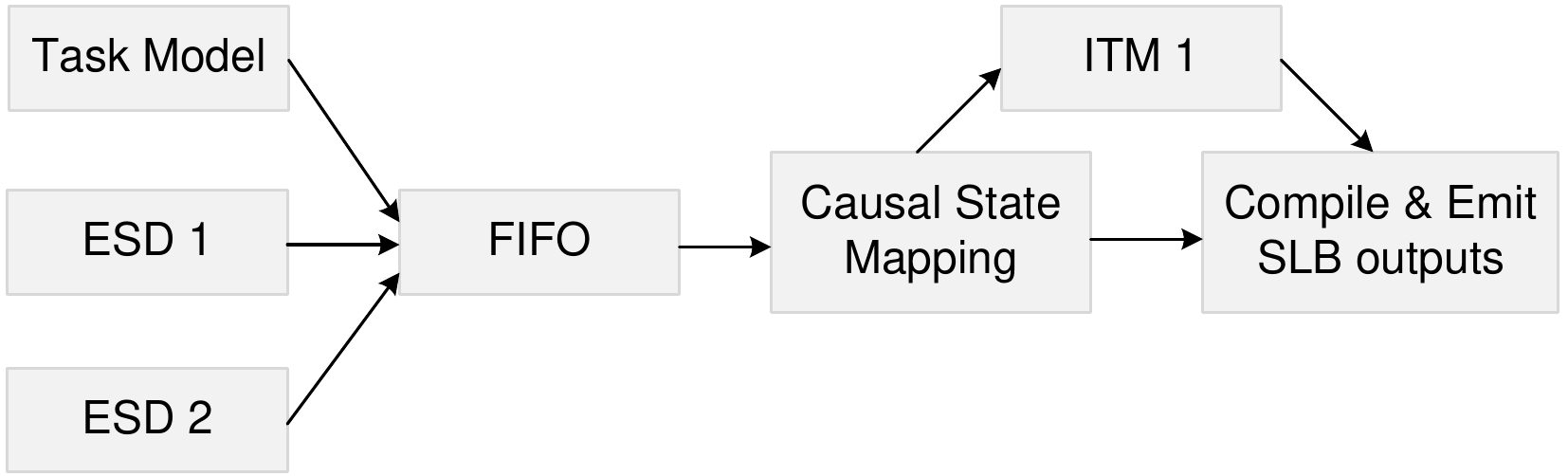}
\caption{Self-labeling modular structures for multiple effects.}
\label{fig_slb_detail}
\end{figure}

\textbf{SLB Implementation.}
A detailed implementation of the self-labeling service is described in \cref{fig_slb_detail}. A non-trivial situation of self-labeling is to cope with multiple asynchronous effects for causal state mapping to self-label the corresponding cause states. The SLB service needs to wait for the delayed effects in order to jointly or individually execute ITMs, or neglects detected effects if no other effects arrive in the given time period and the received effects are unable to determine a unique cause state. We apply a first-in-first-out (FIFO) queue to cache the arrived effects. The Causal State Mapping module regularly scans the FIFO and determines if the current effect states are enough to determine an unambiguous cause state referring to the retrieved causality from KG. In addition, the Causal State Mapping module monitors the timespan and evict effects states that cannot formulate a deterministic cause state due to the lack of necessary effects in a given time period. After the causal state mapping, ITMs are triggered to infer the interaction time by using the assembled effect states as inputs, after which the self-labeling results are compiled and emitted.

\begin{figure}[!t]
\centering
\includegraphics[width=1.0\columnwidth]{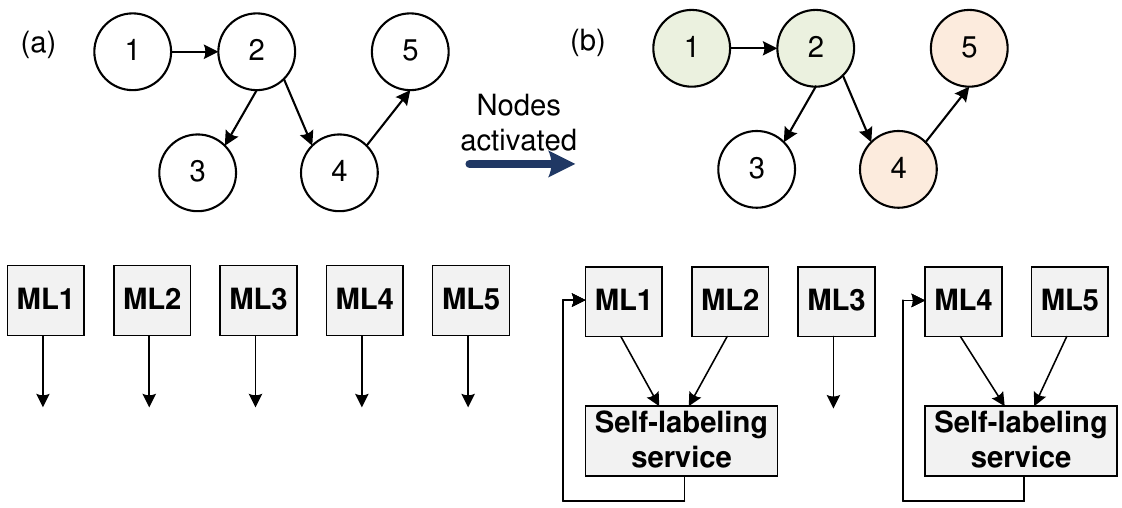}
\caption{An illustration of virtual interactions among ML services due to the initialization of pair-wise self-labeling.}
\label{fig_5nodes}
\end{figure}

\cref{fig_5nodes} illustrates the interactions among ML services with the self-labeling mechanism guided by the causal knowledge graph. Initially, five ML services operate independently for state change detection of the corresponding nodes in the KG. By choosing some of the causally linked nodes for self-labeling as colored in \cref{fig_5nodes}(b), the corresponding ML services start interacting with each other via the defined self-labeling workflow. In turn, self-labeling improves detection accuracy, and the ML services can infer again independently.

\subsection{System Characterization}

A system characterization of several key performance indicators is conducted to evaluate the performance of the proposed AdaptIoT system. The system's backend and frontend applications are deployed on a workstation with a 20-core Intel Xeon W-2155 at 3.30 GHz. The workstation's Ethernet data transfer rate is 1 Gbps. 

We use Raspberry Pi 3B with 1G RAM and 300 Mbps Ethernet as the host processor for multiple time-series sensors installed on machines. Depending on sensor types, the sampling frequency of each sensor ranges from a few hundred Hz to 0.2 Hz. A standard configuration of a sensor node for characterization purposes consists of one host processor, one 6-DOF IMU sensor, one CTH ($CO_{2}$, temperature, humidity) sensor, and one distance (time of flight) sensor, while it can also be customized freely. 
For a single edge node with one IMU, one CTH, and one distance sensor, the average end-to-end timing performance from the data generator to the database is evaluated, and the results are shown in \cref{tb_edgenode}.

\begin{table}[]
    \caption{Test results of a single edge node}
    \resizebox{\columnwidth}{!}{%
    \begin{tabular}{llll}
    \toprule
    Mean throughput & Mean msg size & Mean Delay & Max Delay \\ \midrule
    284 msg/s       & 250.2 byte   & 31 ms      & 64 ms      \\ \bottomrule
    \end{tabular}%
    }
    \label{tb_edgenode}
\end{table}

As for camera streaming, Raspberry Pi 4B with 8G RAM and 1 Gbps Ethernet is chosen. Correspondingly, Raspberry Pi Camera Module 3 with 1080p resolution and 30 fps is used. Each camera produces streams simultaneously in two modes: preview and full HD resolution. The preview mode streams at 240p resolution for GUI display only. The average end-to-end delay is 39 ms. The full HD mode streams at 1080p to the video segmenter for self-labeling and the corresponding ML services for inference. This design ensures that the acquired video dataset and ML inference can use high-quality images while reducing bandwidth requirements for GUI users. The average frame size is 69 KB, and theoretically, using 1 Gbit, the system can support about 60 cameras simultaneously.

To provide a baseline for characterization, we detail the system configuration below. First, a mock test is accomplished to evaluate the maximum capacity of a single Kafka producer and consumer. We use a laptop with an AMD Ryzen 7 6800H 16-core CPU and 1 Gbps network as the transmitter for hosting the mock sensor. An Apache Kafka cluster with 3 nodes and 10 partitions is used. The Kafka single producer test result is shown in \cref{tb_kafka}. The maximum throughput of a single consumer is 388k msg/s, which is equivalent to 92.5 MB/s. 
Note that the test is accomplished with only one producer, one consumer, and three Kafka nodes. Due to the horizontal scalability of Kafka, higher performance can be realized by proper scaling.

Additionally, a realistic system capability test is conducted on the real deployment. We start with 3 standard sensor nodes, 9 additional power meters with 1 Hz data rate, and 3 additional edge sensor services, including a Tiny ML board, an IMU sensor, and a data query service for the UR3e robot, while 8 camera streamings run in the background. In total, 29 edge services are actively running; on average, 108 million messages are generated daily. We monitor the data ingestion speed of InfluxDB, and the average data ingestion rate is 1259 messages per second (msg/s). Based on a single consumer with an average receiving speed of 92.5 MB/sec and a time-series edge producer with an average data rate of 41 msg/s, the theoretically maximum number of supported time-series edge services is 13.2k.

\begin{table}
    \centering
    \caption{Kafka Single Producer Throughput Test Results}
    \resizebox{1.0\columnwidth}{!}{%
    \begin{tabular}{@{}llllll@{}}
    \toprule
     Max number of messages      &  Mean delay  & Max delay \\ \midrule
    182k     & 679 ms & 979 ms \\ \bottomrule
    \end{tabular}%
    }
    \label{tb_kafka}
\end{table}

\section{A real self-labeling experiment running on AdaptIoT}
\label{sec: exp}

To demonstrate the applicability of the proposed system for self-labeling applications,
a self-labeling application is developed and deployed on AdaptIoT to demonstrate the system's efficacy. The self-labeling application utilizes the example in \cite{tii} and replicates the adaptive worker-machine interaction detection on a 3D printer. The experiment is to use the concept of interactive causality to design the self-labeling system for adapting a worker action recognition model. The cause side uses cameras to detect body gestures as an indication of worker-machine interactions. The effect side uses a power meter to detect machine responses in the form of energy consumption. 

The developed self-labeling application is driven by a causal knowledge graph that describes the extracted domain knowledge. This KG representing the causality embedded in the 3D printer operation among people, machines, and materials is built and loaded to the graph database with corresponding metadata as shown in \cref{fig_kg}. As an illustration, we implement five sensors for five nodes in the KG with five corresponding ML services to detect the state change of each node. Among the five implemented nodes, two interactive nodes are chosen to conduct self-labeling, as highlighted in the red circle. The two nodes represent a worker's body movement and the machine's status change.

\begin{figure}[!t]
\centering
\includegraphics[width=1.0\columnwidth]{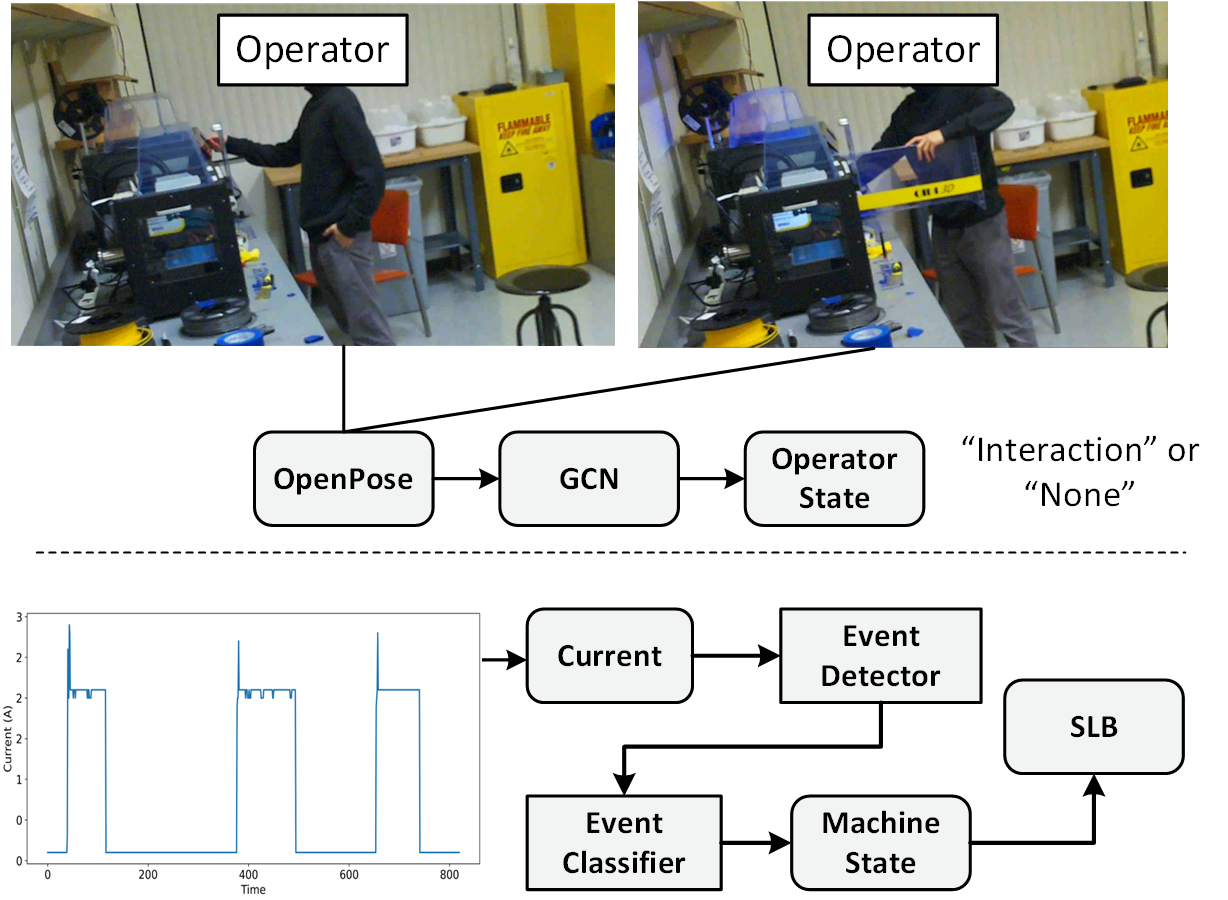}
\caption{Experimental setup of the 3D printer self-labeling use case. (a) illustrates the task model data processing pipeline, and (b) describes the ESD pipeline for current signals.}
\label{fig_exp}
\end{figure}

The implementation details are shown in \cref{fig_exp}. The worker action is defined as binary, namely interaction and non-interaction. The interaction is defined as when workers push the power button to turn on/off the 3D printer, which corresponds to a change in the machine's power consumption as an effect. An ML service composed of a cascaded OpenPose \cite{cao2017realtime} and graph convolution network (GCN) \cite{tii} is implemented to recognize worker action as the task model. To detect a machine's power change, a machine state recognition algorithm composed of an event detector and classifier is implemented as the ESD \cite{jms}. The ITM uses a lookup table and statistical Gaussian model to infer the interaction time. This entire self-labeling application runs on the proposed AdaptIoT system.

To demonstrate effectiveness, we manually collected and labeled a dataset of 400 samples as the validation and test sets. Through the experiment, a self-labeled dataset composed of 200 samples is automatically collected and labeled using the AdaptIoT system over three weeks of 3D printer usages. \cref{tb_exp} summarizes the accuracy compared with several semi-supervised approaches. The results show the mean and standard deviation derived over the training with 10 random seeds. By default, all other semi-supervised approaches apply the temporal random shift as the data augmentation. It can be observed that the self-labeling method consistently outperforms other semi-supervised methods with a smaller standard deviation, indicating a more stabilized training, which demonstrates the applicability of the proposed AdaptIoT system for the self-labeling applications. 
According to the theory in \cite{ren2023slb}, the self-labeling and semi-supervised methods show comparable performance when there is no observable data distribution shift as in the situation of this experiment. The merit of self-labeling over traditional semi-supervised methods mainly manifests in the scenario of data distribution shifts, which has been demonstrated by previous studies and is not the scope of this study.

\begin{table}
    \centering
    \caption{Model Accuracy (\%) trained on the experiment dataset}
    \resizebox{0.6\columnwidth}{!}{%
    \begin{tabular}{@{}llllll@{}}
    \toprule
     Method       &  Accuracy   \\ \midrule
    PseudoLabel \cite{2013pseudo} & 90.05 ($\pm 2.72$)  \\
    SimMatch \cite{simmatch} & 96.18 ($\pm 1.48$)  \\ 
    SoftMatch \cite{chen2023softmatch} & 98.22 ($\pm 0.68$)  \\ 
    FreeMatch \cite{wang2023freematch} & 97.90 ($\pm 0.72$)  \\ \midrule
    Pretrain only  & 95.21 ($\pm 2.72$)  \\
    SLB (w/ data aug.)    & 98.24 ($\pm 0.57$)  \\ 
    SLB (w/o data aug.)     & 98.33 ($\pm 0.22$)  \\ \bottomrule
    \end{tabular}%
    }
    \label{tb_exp}
\end{table}

An interesting observation is found from the experiment results that the temporal random shift as the data augmentation adversely affect the self-labeling accuracy.
\cite{ren2024book} proposes a qualitative explanation of the impact of the uncertain interaction time on the self-labeling and model retraining performance. They use the concept of motion smoothness to explain that even though the ITM may infer the interaction time at a deviated timestamp, the natural motion smoothness alleviates the adverse effect of deviated interaction time. Hypothetically, the inaccuracy caused by interaction time inference is equivalent to the temporal random shift. The adverse effect of adding the temporal random shift to the self-labeling shown in \cref{tb_exp} partially reveals this fact but requires deeper research in the future.

\section{Conclusion}
\label{sec: end}
In this study, an IoT system, AdaptIoT, is designed and demonstrated to support the interactive causality-enabled self-labeling workflow for developing adaptive machine learning applications in cyber manufacturing. The AdaptIoT is designed as a web-based and microservice platform for both manufacturing IoT digitization and intelligentization with an end-to-end data streaming component, a machine learning integration component, and a self-labeling service. AdaptIoT ensures high throughput and low latency data acquisition and seamless integration and deployment of ML applications.
The self-labeling service automates the entire self-labeling workflow to allow real-time and parallel task model adaptation. 
A university laboratory as a makespace is retrofit with the AdaptIoT system for future adaptive learning cyber manufacturing application development. Overall, more adaptive ML applications in cyber manufacturing are envisioned to be developed in the future based on the proposed AdaptIoT system.

\section*{Acknowledgment}
The authors would like to thank Broadcom Foundation and Calit2 for the funding support.

\ifCLASSOPTIONcaptionsoff
  \newpage
\fi

\bibliographystyle{IEEEtran}
\bibliography{IEEEabrv,Bibliography}

\begin{thebibliography}{10}
\providecommand{\url}[1]{#1}
\csname url@samestyle\endcsname
\providecommand{\newblock}{\relax}
\providecommand{\bibinfo}[2]{#2}
\providecommand{\BIBentrySTDinterwordspacing}{\spaceskip=0pt\relax}
\providecommand{\BIBentryALTinterwordstretchfactor}{4}
\providecommand{\BIBentryALTinterwordspacing}{\spaceskip=\fontdimen2\font plus
\BIBentryALTinterwordstretchfactor\fontdimen3\font minus \fontdimen4\font\relax}
\providecommand{\BIBforeignlanguage}[2]{{%
\expandafter\ifx\csname l@#1\endcsname\relax
\typeout{** WARNING: IEEEtran.bst: No hyphenation pattern has been}%
\typeout{** loaded for the language `#1'. Using the pattern for}%
\typeout{** the default language instead.}%
\else
\language=\csname l@#1\endcsname
\fi
#2}}
\providecommand{\BIBdecl}{\relax}
\BIBdecl

\bibitem{lu2016internet}
Y.~Lu and J.~Cecil, ``An internet of things (iot)-based collaborative framework for advanced manufacturing,'' \emph{The International Journal of Advanced Manufacturing Technology}, vol.~84, pp. 1141--1152, 2016.

\bibitem{ser_ori1}
F.~Tao and Q.~Qi, ``New it driven service-oriented smart manufacturing: Framework and characteristics,'' \emph{IEEE Transactions on Systems, Man, and Cybernetics: Systems}, vol.~49, no.~1, pp. 81--91, 2019.

\bibitem{cps_microservice}
K.~Thramboulidis, D.~C. Vachtsevanou, and I.~Kontou, ``Cpus-iot: A cyber-physical microservice and iot-based framework for manufacturing assembly systems,'' \emph{Annual Reviews in Control}, vol.~47, pp. 237--248, 2019.

\bibitem{rudack2022towards}
M.~Rudack, M.~Rath, U.~Vroomen, and A.~B{\"u}hrig-Polaczek, ``Towards a data lake for high pressure die casting,'' \emph{Metals}, vol.~12, no.~2, p. 349, 2022.

\bibitem{yenSaaS}
I.-L. Yen, S.~Zhang, F.~Bastani, and Y.~Zhang, ``A framework for iot-based monitoring and diagnosis of manufacturing systems,'' in \emph{2017 IEEE Symposium on Service-Oriented System Engineering (SOSE)}, 2017, pp. 1--8.

\bibitem{MOURTZIS2016290}
D.~Mourtzis, E.~Vlachou, and N.~Milas, ``Industrial big data as a result of iot adoption in manufacturing,'' \emph{Procedia CIRP}, vol.~55, pp. 290--295, 2016, 5th CIRP Global Web Conference - Research and Innovation for Future Production (CIRPe 2016).

\bibitem{LIU2022102217}
C.~Liu, Z.~Su, X.~Xu, and Y.~Lu, ``Service-oriented industrial internet of things gateway for cloud manufacturing,'' \emph{Robotics and Computer-Integrated Manufacturing}, vol.~73, p. 102217, 2022.

\bibitem{SHENG2024102324}
Y.~Sheng, G.~Zhang, Y.~Zhang, M.~Luo, Y.~Pang, and Q.~Wang, ``A multimodal data sensing and feature learning-based self-adaptive hybrid approach for machining quality prediction,'' \emph{Advanced Engineering Informatics}, vol.~59, p. 102324, 2024.

\bibitem{MORARIU2020103244}
C.~Morariu, O.~Morariu, S.~Răileanu, and T.~Borangiu, ``Machine learning for predictive scheduling and resource allocation in large scale manufacturing systems,'' \emph{Computers in Industry}, vol. 120, p. 103244, 2020.

\bibitem{ml_deploy_challenge}
A.~Paleyes, R.-G. Urma, and N.~D. Lawrence, ``Challenges in deploying machine learning: A survey of case studies,'' \emph{ACM Comput. Surv.}, vol.~55, no.~6, dec 2022.

\bibitem{davis2020cyberinfrastructure}
J.~Davis, H.~Malkani, J.~Dyck, P.~Korambath, and J.~Wise, ``Cyberinfrastructure for the democratization of smart manufacturing,'' in \emph{Smart Manufacturing}.\hskip 1em plus 0.5em minus 0.4em\relax Elsevier, 2020, pp. 83--116.

\bibitem{yan2021augmented}
H.~Yan, Y.~Guo, and C.~Yang, ``Augmented self-labeling for source-free unsupervised domain adaptation,'' in \emph{NeurIPS 2021 Workshop on Distribution Shifts: Connecting Methods and Applications}, 2021.

\bibitem{slb_neurips}
P.~Zhou, C.~Xiong, X.~Yuan, and S.~C.~H. Hoi, ``A theory-driven self-labeling refinement method for contrastive representation learning,'' in \emph{Advances in Neural Information Processing Systems}, vol.~34, 2021, pp. 6183--6197.

\bibitem{delay2}
M.~Grzenda, H.~M. Gomes, and A.~Bifet, ``Performance measures for evolving predictions under delayed labelling classification,'' in \emph{2020 International Joint Conference on Neural Networks (IJCNN)}, 2020, pp. 1--8.

\bibitem{delayed_lb_review}
H.~M. Gomes, M.~Grzenda, R.~Mello, J.~Read, M.~H. Le~Nguyen, and A.~Bifet, ``A survey on semi-supervised learning for delayed partially labelled data streams,'' \emph{ACM Comput. Surv.}, feb 2022.

\bibitem{domainknowledge}
A.~J. Ratner, C.~M. De~Sa, S.~Wu, D.~Selsam, and C.~R\'{e}, ``Data programming: Creating large training sets, quickly,'' in \emph{Advances in Neural Information Processing Systems}, vol.~29, 2016.

\bibitem{stewart2017label}
R.~Stewart and S.~Ermon, ``Label-free supervision of neural networks with physics and domain knowledge,'' in \emph{Thirty-First AAAI Conference on Artificial Intelligence}, 2017.

\bibitem{ren2023slb}
Y.~Ren, A.~H. Yen, and G.-P. Li, ``A self-labeling method for adaptive machine learning by interactive causality,'' \emph{IEEE Transactions on Artificial Intelligence}, 2023.

\bibitem{tii}
Y.~Ren and G.-P. Li, ``An interactive and adaptive learning cyber physical human system for manufacturing with a case study in worker machine interactions,'' \emph{IEEE Transactions on Industrial Informatics}, vol.~18, no.~10, pp. 6723--6732, 2022.

\bibitem{conceptdrift}
J.~Lu, A.~Liu, F.~Dong, F.~Gu, J.~Gama, and G.~Zhang, ``Learning under concept drift: A review,'' \emph{IEEE transactions on knowledge and data engineering}, vol.~31, no.~12, pp. 2346--2363, 2018.

\bibitem{ckg}
H.~Huang, ``Causal relationship over knowledge graphs,'' in \emph{Proceedings of the 31st ACM International Conference on Information \& Knowledge Management}, ser. CIKM '22.\hskip 1em plus 0.5em minus 0.4em\relax New York, NY, USA: Association for Computing Machinery, 2022, p. 5116–5119.

\bibitem{timelag}
H.~F. Gollob and C.~S. Reichardt, ``Taking account of time lags in causal models,'' \emph{Child Development}, vol.~58, no.~1, pp. 80--92, 1987.

\bibitem{pearl2009causality}
J.~Pearl, \emph{Causality}, ser. Causality: Models, Reasoning, and Inference.\hskip 1em plus 0.5em minus 0.4em\relax Cambridge University Press, 2009.

\bibitem{mlops}
D.~Kreuzberger, N.~K{\"u}hl, and S.~Hirschl, ``Machine learning operations (mlops): Overview, definition, and architecture,'' \emph{IEEE access}, 2023.

\bibitem{merkel2014docker}
D.~Merkel \emph{et~al.}, ``Docker: lightweight linux containers for consistent development and deployment,'' \emph{Linux j}, vol. 239, no.~2, p.~2, 2014.

\bibitem{dobbelaere2017kafka}
P.~Dobbelaere and K.~S. Esmaili, ``Kafka versus rabbitmq: A comparative study of two industry reference publish/subscribe implementations: Industry paper,'' in \emph{Proceedings of the 11th ACM international conference on distributed and event-based systems}, 2017, pp. 227--238.

\bibitem{mq2}
G.~Fu, Y.~Zhang, and G.~Yu, ``A fair comparison of message queuing systems,'' \emph{IEEE Access}, vol.~9, pp. 421--432, 2020.

\bibitem{influxdb}
S.~Rinaldi, F.~Bonafini, P.~Ferrari, A.~Flammini, E.~Sisinni, and D.~Bianchini, ``Impact of data model on performance of time series database for internet of things applications,'' in \emph{2019 IEEE International Instrumentation and Measurement Technology Conference (I2MTC)}.\hskip 1em plus 0.5em minus 0.4em\relax IEEE, 2019, pp. 1--6.

\bibitem{neo4j}
D.~Fernandes, J.~Bernardino \emph{et~al.}, ``Graph databases comparison: Allegrograph, arangodb, infinitegraph, neo4j, and orientdb.'' \emph{Data}, vol.~10, 2018.

\bibitem{sse}
S.~Vinoski, ``Server-sent events with yaws,'' \emph{IEEE internet computing}, vol.~16, no.~5, pp. 98--102, 2012.

\bibitem{covington2016deep}
P.~Covington, J.~Adams, and E.~Sargin, ``Deep neural networks for youtube recommendations,'' in \emph{Proceedings of the 10th ACM conference on recommender systems}, 2016, pp. 191--198.

\bibitem{cao2017realtime}
Z.~Cao, T.~Simon, S.-E. Wei, and Y.~Sheikh, ``Realtime multi-person 2d pose estimation using part affinity fields,'' in \emph{CVPR}, 2017.

\bibitem{jms}
Y.~Ren and G.~P. Li, ``A contextual sensor system for non-intrusive machine status and energy monitoring,'' \emph{Journal of Manufacturing Systems}, vol.~62, pp. 87--101, 2022.

\bibitem{2013pseudo}
D.-H. Lee \emph{et~al.}, ``Pseudo-label: The simple and efficient semi-supervised learning method for deep neural networks,'' in \emph{Workshop on challenges in representation learning, ICML}, vol.~3, no.~2, 2013, p. 896.

\bibitem{simmatch}
M.~Zheng, S.~You, L.~Huang, F.~Wang, C.~Qian, and C.~Xu, ``Simmatch: Semi-supervised learning with similarity matching,'' in \emph{Proceedings of the IEEE/CVF Conference on Computer Vision and Pattern Recognition (CVPR)}, June 2022, pp. 14\,471--14\,481.

\bibitem{chen2023softmatch}
H.~Chen, R.~Tao, Y.~Fan, Y.~Wang, J.~Wang, B.~Schiele, X.~Xie, B.~Raj, and M.~Savvides, ``Softmatch: Addressing the quantity-quality trade-off in semi-supervised learning,'' 2023.

\bibitem{wang2023freematch}
Y.~Wang, H.~Chen, Q.~Heng, W.~Hou, Y.~Fan, , Z.~Wu, J.~Wang, M.~Savvides, T.~Shinozaki, B.~Raj, B.~Schiele, and X.~Xie, ``Freematch: Self-adaptive thresholding for semi-supervised learning,'' 2023.

\bibitem{ren2024book}
Y.~Ren, A.~Yen, S.~Saraj, and G.~P. Li, ``Interactive causality-enabled adaptive machine learning in cyber-physical systems: Technology and applications in manufacturing and beyond,'' in \emph{Principles and Applications of Adaptive Artificial Intelligence}.\hskip 1em plus 0.5em minus 0.4em\relax IGI Global, 2024, pp. 179--206.

\end{thebibliography}

\end{document}